\documentclass{article}
\usepackage[dvips,pdftex]{graphicx}
\DeclareGraphicsExtensions{.pdf,.jpeg,,jpg,.png}
\graphicspath{{./figs/}}

\usepackage{algorithm}
\usepackage{algpseudocode}
\usepackage{varwidth}
\usepackage{listings}
\usepackage[OT4,T1]{fontenc}
\usepackage[cmex10]{amsmath}
\usepackage{url}
\usepackage{multirow}
\usepackage{todonotes}
\usepackage{rotating}

\begin{document}
\label{firstpage}

\title
      {Multi-swarm PSO algorithm for the Quadratic Assignment Problem: a massive parallel implementation on the OpenCL platform}

\author
       {Piotr Szwed and Wojciech Chmiel\\
AGH University of Science and Technology\\
Al. Mickiewicza 30\\
30\,059 Krak\'ow, Poland\\
\{pszwed@agh.edu.pl, wch@agh.edu.pl\}
}

\date{\today}

%
%

\maketitle

\begin{abstract}
This paper presents a multi-swarm PSO algorithm for the Quadratic Assignment Problem (QAP) implemented on OpenCL platform. Our work was motivated by results of time efficiency tests performed for single-swarm algorithm implementation that showed clearly that  the benefits of a parallel execution platform can be fully exploited, if the processed population is large.
The described algorithm can be executed in two modes: with independent swarms or with migration. We discuss the algorithm construction, as well as we report  results of tests performed on several problem instances from the QAPLIB library.
During the experiments the algorithm was configured to process large populations. This allowed us to collect statistical data related to values of goal function reached by individual particles. We use them to demonstrate on two test cases that although single particles seem to behave chaotically during the optimization process, when the whole population is analyzed, the probability that a particle will select a near-optimal solution grows.

\end{abstract}

%

\section{Introduction}
\label{sec:intro}

Quadratic Assignment Problem (QAP) \cite{KBe57,Cela98} is a well known combinatorial  problem that can be used as optimization model in many areas \cite{stickel:Bermudez01,MasonR97,Groetschel1991}.    
QAP may be formulated as follows: given a set of $n$ \emph{facilities} and $n$ \emph{locations}, the goal is to find an assignment of facilities to unique locations that minimizes the  sum of flows between facilities multiplied by distances between their locations. As the problem is NP hard \cite{Sahni76}, it can be solved optimally for small problem instances. For larger problems (n>30), several heuristic algorithm were proposed \cite{Taillard199587,Misevicius12,ChK11}. 

One of the discussed methods \cite{onwubolu2004particle,Liu2007} is  the Particle Swarm Optimization (PSO).  It attempts to find an optimal problem solution by moving a population of particles in the search space. Each particle is characterized by two features its position and velocity. Depending on a method variation, particles may exchange information on their positions and reached values of goal functions \cite{EberhartKennedy95}.

In our recent work \cite{SzChKad2015} we have developed PSO algorithm for the Quadratic Assignment Problem on OpenCL platform. The algorithm was capable of processing one swarm, in which particles shared information about the global best solution to update their search directions. Following typical patterns for GPU based calculations, the implementation was a combination of parallel tasks (kernels) executed on GPU orchestrated by sequential operations run on the host (CPU). Such organization of computations involves inevitable  overhead related to  data transfer between the host and the GPU device. The time efficiency test reported in \cite{SzChKad2015} showed clearly that the benefits of a parallel execution platform can be fully exploited, if processed populations are large, e.g. if they comprise several hundreds or thousands particles. For smaller populations sequential algorithm implementation was superior both as regards the total swarm processing time and the  time required to process one particle. This suggested a natural improvement of the previously developed algorithm: by scaling it up to high numbers of particles organized into several swarms.

In this paper we discuss a multi-swarm implementation PSO algorithm for the QAP problem on OpenCL platform. The algorithm can be executed in two modes: with independent swarms, each maintaining its best solution, or with migration between swarms. We describe the algorithm construction, as well as we report tests performed on several problem instances from the QAPLIB library \cite{QAPLIB}. 
Their results show advantages of massive parallel computing: the obtained solutions are very close to optimal or best known for particular problem instances.  

The developed algorithm is not designed to exploit the problem specificity (see for example \cite{Fischetti2012}), as well as it is not intended to compete with supercomputer or grid based implementations providing exact solutions for the QAP \cite{anstreicher2002solving}. On the contrary, we are targeting low-end GPU devices, which are present in most laptops and workstations in everyday use, and accept near-optimal solutions. 

During the tests the algorithm was configured to process large numbers of particles (in the order of 10000). This allowed us to collect  data related to goal function values reached by individual particles and present such statistical measures as percentile ranks and probability mass functions for the whole populations or selected swarms. 

%

The paper is organized as follows: next Section~\ref{sec:related-works} discusses the QAP problem, as well as the PSO method.  It is followed by Section~\ref{sec:methods}, which describes the adaptation of PSO to the QAP and the  parallel implementation on the OpenCL platform. Experiments performed and their results are presented in Section~\ref{sec:experiments-results}.  Section~\ref{sec:conclusions} provides concluding remarks.

\section{Related works}
\label{sec:related-works}

\subsection{Quadratic Assignment Problem}
\label{subsec:QAP}

Quadratic Assignment Problem was introduced by Koopmans and Beckman in 1957 as a mathematical model describing assignment of economic activities to a set of locations \cite{KBe57}. 

Let $V=\left \{1,...,n \right \}$ be a set of \emph{locations} (nodes) linked by $n^2$ arcs. Each arc linking a pair of nodes $(k,l)$ is attributed with a non-negative weight $d_{kl}$ interpreted as a distance. Distances are usually presented in form of  $n \times n$ distance matrix  $D=\left[d_{kl}\right]$. The next problem component is a set of  
facilities $N=\left \{1,...,n \right \}$ and a $n \times n$ non-negative flow matrix $F=\left[f_{ij}\right]$, whose elements describe flows between pairs of facilities $(i,j)$.

The problem goal is to find an assignment $\pi \colon N \to V$ that minimizes the total cost calculated as sum of flows $f_{ij}$ between pairs of facilities $(i,j)$ multiplied by distances $d_{\pi(i)\pi(j)}$ between pairs of locations $(\pi(i),\pi(j))$, to which they are assigned. The permutation $\pi$ can be encoded as $n^2$ binary variables $x_{ki}$, where $k=\pi(i)$, what gives the following problem statement:          
    
\begin{equation}
\label{eq:qap}
min\,\sum_{i=1}^{n}\,\sum_{j=1}^{n}\,\sum_{k=1}^{n}\,\sum_{l=1}^{n}\,f_{ij}d_{kl}x_{ki}x_{lj}
\end{equation}

subject to:
\begin{equation}
\label{eq:permutation-matrix}
\renewcommand{\arraystretch}{1.5}
\begin{array}{lcr}
\sum_{i=1}^n x_{ij}= 1,&&\text{for } 1\leq j \leq n\\
\sum_{j=1}^n x_{ij}= 1,&&\text{for } 1\leq i \leq n\\
x_{ij}\in \{0,1\}
\end{array}
\end{equation}



The $n\times n$ matrix $X = [x_{ki}]$ satisfying (\ref{eq:permutation-matrix}) is called permutation matrix.
 
In most cases matrix $D$ and $F$ are symmetric. Moreover, their diagonal elements are often equal 0. Otherwise, the component $f_{ii} d_{kk} x_{ki} x_{ki}$ can be extracted as a linear part of the goal function interpreted as an  installation cost of $i$-th facility  at $k$-th location .

QAP models found application in various areas including transportation \cite{stickel:Bermudez01}, scheduling, electronics (wiring problem), distributed computing, statistical data analysis (reconstruction of destroyed soundtracks), balancing of turbine running \cite{MasonR97}, chemistry 
, genetics \cite{Phillips94aquadratic}, creating the control panels and manufacturing~\cite{Groetschel1991}.


In 1976 Sahni and Gonzalez proved that the QAP is strongly $\mathcal{ NP}$\emph{-hard} \cite{Sahni76}, by showing that a hypothetical existence of a polynomial time algorithm for solving the QAP would imply an existence of a polynomial time algorithm for an $\mathcal{NP}$\emph{-complete} decision problem - the Hamiltonian cycle.    


In many research works QAP is considered one of the most challenging optimization problem. This in particular regards problem instances gathered in a publicly available and continuously updated QAPLIB library  \cite{QAPLIB,BKR91}. A practical size limit for problems that can be solved with exact algorithms  is about $n=30$ \cite{hahn2010exact}. In many cases optimal solutions were found  with branch and bound algorithm requiring high computational power offered by computational grids \cite{anstreicher2002solving} or supercomputing clusters equipped with a few dozen of processor cores and hundreds gigabytes of memory \cite{hahn2013memory}. On the other hand, in \cite{Fischetti2012} a very successful approach exploiting the problem structure was reported. It allowed to solve several hard problems from QAPLIB using very little resources. 

A number of heuristic algorithms allowing to find a near-optimal solutions for QAP were proposed.  They include  Genetic Algorithm \cite{ahuja2000greedy}, various versions of Tabu search \cite{Taillard199587}, Ant Colonies 
\cite{stutzle1999aco,gambardella1999ant} and Bees algorithm \cite{fon2010investigating}. Another method, being discussed further, is Particle Swarm Optimization \cite{onwubolu2004particle,Liu2007} .


%

\subsection{Particle Swarm Optimization}
\label{subsec:related-work-pso}

The classical PSO algorithm \cite{EberhartKennedy95} is an optimization method defined for continuous domain.    
During the optimization process a number of particles move through a search space and update their state  at discrete time steps $t=1,2,3,\dots$ Each particle is characterized by position $x(t)$ and velocity $v(t)$. A particle remembers its best position reached so far $p^L(t)$, as well as it can use information about the best solution found by the swarm $p^G(t)$. 

The state equation for a particle is given by the formula (\ref{eq:pso-state-eq}). Coefficients $c_1, c_2, c_3 \in[0,1]$ are called respectively \emph{inertia}, \emph{cognition} (or \emph{self recognition})  and \emph{social} factors, whereas $r_1,r_2$ are random numbers uniformly distributed in $[0,1]$   


\begin{align}
\label{eq:pso-state-eq}
v(t+1) &= c_1\cdot v(t) + c_2 \cdot r_2(t)\cdot(p^L(t) - x(t))
\left.{} + c_3\cdot r_3(t) \cdot(p^G(t) - x(t)) \right.\nonumber\\
x(t+1) &= x(t) + v(t)
\end{align}

An adaptation of the PSO method to a discrete domain necessities in giving interpretation to the velocity concept, as well as defining equivalents of scalar multiplication, subtraction and addition for arguments being solutions and velocities. Examples of such interpretations can be found in \cite{clerc2004discrete} for the TSP and \cite{onwubolu2004particle} for the QAP. 

A PSO algorithm for solving QAP using similar representations of particle state was proposed by Liu et al. \cite{Liu2007}. Although the approach presented there was inspiring, the paper gives very little information on efficiency of the developed algorithm.

\subsection{GPU based calculations}
Recently many computationally demanding applications has been redesigned to exploit the capabilities  offered by massively parallel computing GPU platforms. They include such tasks as: physically based simulations, signal processing, ray tracing, geometric computing and data mining \cite{owens2007survey}. Several attempts have been also made to develop various population based optimization algorithms on GPUs  including: the particle swarm optimization  \cite{zhou2009gpu}, the ant colony optimization \cite{acoTabuGPU2013}, the genetic  \cite{geneticGPU2013} and memetic algorithm \cite{memeticGPU2013}. The described implementations benefit from capabilities offered by GPUs by  processing whole populations by fast GPU cores running in parallel.


\section{Algorithm design and implementation} 
\label{sec:methods}

In this section we describe the algorithm design, in particular the adaptation of Particle Swarm Optimization  metaheuristic to the QAP problem, as well as a specific algorithm implementation on OpenCL platform. 
As it was stated in Section~\ref{subsec:related-work-pso}, the PSO uses generic concepts of position $x$ and velocity $v$ that can be mapped to a particular problem in various ways.
Designing an algorithm for a GPU platform requires decisions on how to divide it into parts that are either executed sequentially at the host side or in parallel on the device.

\subsection{PSO adaptation for the QAP problem}

A state of a particle is a pair $(X,V)$. 
In the presented approach both are $n\times n$ matrices, where $n$ is the problem size. The permutation matrix $X=[x_{ij}]$ encodes an assignment of facilities to locations. Its elements $x_{ij}$ are equal to $1$, if $j$-th facility is assigned to $i$-th location, and take value $0$ otherwise. 

A particle moves in the solution space following the direction given by the velocity $V$. Elements $v_{ij}$  have the following interpretation: if $v_{ij}$ has high positive value, then a procedure determining the next solution should favor an assignment $x_{ij}=1$. On the other hand, if $v_{ij}\leq 0$, then $x_{ij}=0$ should be preferred.       

The state of a particle reached in the $t$-th iteration will be denoted by $(X(t),V(t))$.
In each iteration it is updated according to formulas (\ref{eq:velocity}) and (\ref{eq:position}).



\begin{align}
\label{eq:velocity}
V(t+1)&=S_v\left(c_1\cdot V(t)+c_2\cdot r_2(t)\cdot(P^{L}(t)-X(t))+\right.\nonumber\\
&\qquad \left. {} c_3 \cdot r_3(t)\cdot(P^G(t)-X(t))\right)
\end{align}

\begin{equation}
\label{eq:position}
X(t+1)=S_x(X(t)+V(t))
\end{equation}     

Coefficients $r_2$ and $r_3$  are random numbers from $[0,1]$ generated in each iteration for every particle separately. They are introduced to model a random choice between movements in the previous direction (according to $c_1$ -- inertia), the best local solution (self recognition) or the global best solution (social behavior).    

All operators appearing in (\ref{eq:velocity}) and (\ref{eq:position}) are standard operators from the linear algebra. Instead of redefining them for a particular problem,  see e.g. \cite{clerc2004discrete}, we propose to use aggregation functions $S_v$ and $S_x$ that allow to adapt the algorithm to particular needs of a discrete problem.  
%
%
%
%
    
The function $S_v$ is used to assure that velocities have reasonable values. Initially, we thought that unconstrained growth of velocity can be a problem, therefore we have implemented a function, which restricts the elements of $V$ to an interval $[-v_{max},v_{max}]$. This function is referred as \emph{raw} in Table~\ref{tab:optimization-results}. However, the experiments conducted showed, that in case of small inertia factor, e.g. $c_1=0.5$, after a few iterations all velocities tend to 0 and in consequence all particles converge to the best solution encountered earlier by the swarm. To avoid such effect another function that additionally performs column normalization was proposed. For each $j$-th column the sum of absolute values of the elements $n_j = \sum_{i=1}^n |v_{ij}|$ is calculated and then the following assignment is made: $v_{ij}\gets v_{ij}/n_j$. 

%
%
According to formula (\ref{eq:position}) a new  particle position $X(t+1)$ is obtained by aggregating the previous state components: $X(t)$ and $V(t)$. As elements of a matrix $X(t)+V(t)$ may take values from $[-v_{max},v_{max}+1]$, the $S_x$ function is responsible for converting it into a valid permutation matrix satisfying (\ref{eq:permutation-matrix}), i.e. having exactly one $1$ in each row and column. Actually, $S_v$ is rather a procedure, than a function, as it incorporates some elements of random choice. 

Three variants of $S_x$ procedures were implemented:

\begin{enumerate}
\item $GlobalMax(X)$ -- iteratively searches for $x_{rc}$, a maximum  element in a matrix $X$, sets it to $1$ and clears other elements in the row $r$ and $c$. 
\item $PickColumn(X)$ -- picks a column $c$ from $X$, selects a maximum element $x_{rc}$, replaces it by $1$ and clears other elements in $r$ and $c$. 
\item $SecondTarget(X,Z,d)$ -- similar to $GlobalMax(X)$, however during the first $d$ iterations ignores elements $x_{ij}$, such that $z_{ij}=1$. (As the parameter $Z$ a solution $X$ from the last iteration  is used.)  
\end{enumerate}


In several experiments, in which  $GlobalMax$ aggregation procedure was applied, particles seemed to get stuck, even if their velocities were far from zero \cite{SzChKad2015}. We reproduce this effect on a small $3 \times 3$ example:

{
\centering
\vspace{12pt}

\begin{tabular}{rr}

$X=
  \left[ {\begin{array}{ccc}
   1 & 0 & 0 \\
   0 & 0 & 1 \\
   0 & 1 & 0 \\
  \end{array} } \right]$
&$V=
  \left[ {\begin{array}{ccc}
   7 & 1 & 3 \\
   0 & 4 & 5 \\
   2 & 3 & 2 \\
  \end{array} } \right]$\\ 
\vspace{-8pt}\\
$ 
X+V=  \left[ {\begin{array}{ccc}
   8 & 1 & 3 \\
   0 & 4 & 6 \\
   2 & 4 & 2 \\
  \end{array} } \right]
$  
&$S_x(X+V) = 
  \left[ {\begin{array}{ccc}
   1 & 0 & 0 \\
   0 & 0 & 1 \\
   0 & 1 & 0 \\
  \end{array} } \right]
$
\end{tabular}

\vspace{12pt}
}
If $GlobalMax(X(t)+V(t))$ is used for the described case, in subsequent iterations it will hold: $X(t+1)=X(t)$, until another particle is capable of changing $(P^G(t)-X(t)))$ component of formula (\ref{eq:velocity}) for velocity calculation. 

A solution for this problem can be to move a particle to a secondary direction (target), by ignoring $d<n$ elements that are in the solution $X(t)$ already set to $1$. This, depending on $d$,  gives an opportunity to reach other solutions, hopefully yielding smaller goal function values.  If chosen elements are maximal  in the remaining matrix, denoted here as $X\oslash_d V$, they are still reasonable movement directions.  
For the discussed example possible values of $X\oslash_d V$ matrices, where $d=1$ and $d=2$, are shown below (\ref{eq:sec-targ}). Elements of a new solution are marked with circles, whereas the upper index indicates the iteration, in which the element was chosen.       

\begin{align}
\label{eq:sec-targ}
X\oslash_{1} V=  
\left[ 
  \begin{array}{ccc}
   \mbox{\textcircled{8}}^2 & 1 & 3 \\
   0 & 4 & \mbox{\textcircled{6}}^1 \\
   2 & \mbox{\textcircled{4}}^3 & 2 \\
  \end{array}  
\right] \nonumber\\
X\oslash_{2} V=  
\left[ 
  \begin{array}{ccc}
   8 & 1 & \mbox{\textcircled{3}}^2 \\
   \mbox{\textcircled{0}}^3 & 4 & 6 \\
   2 & \mbox{\textcircled{4}}^1 & 2 \\
  \end{array}  
\right] \nonumber\\ \text{ or } \nonumber\\
X\oslash_{2} V=  
\left[ 
  \begin{array}{ccc}
   8 & 1 & \mbox{\textcircled{3}}^2 \\
   0 & \mbox{\textcircled{4}}^1 & 6 \\
   \mbox{\textcircled{2}}^3 & 4 & 2 \\
  \end{array}  
\right]
\end{align}

It can be observed that for $d=1$ the value is exactly the same, as it would result from the $GlobalMax$, however setting $d=2$ allows to reach a different solution. The pseudocode of  $SecondTarget$ procedure is listed in Algorithm~\ref{alg:secondtarget}. 




\begin{algorithm}
\caption{Aggregation procedure $SecondTarget$ }
\label{alg:secondtarget}
\begin{algorithmic}[1]
\Require $X = Z+V$ - new solution requiring normalization  
\Require $Z$ - previous solution
\Require $depth$ - number of iterations, in which during selection of maximum element the algorithm ignore positions, where corresponding element of $Z$ is equal~to~1    

\Procedure{SecondTarget}{$X$,$Z$,$depth$}
    
    \State $R \gets \{1,\dots,n\}$ 
    \State $C \gets\{1,\dots,n\}$ 
    \For {$i$ in (1,n)}
    \State   Calculate $M$, the set of maximum elements
    \If{$i\leq depth$} 
	\State Ignore elements $x_{ij}$ such that $z_{ij}=1$
    \State $M \gets$ \par \hskip 1cm $ \{(r,c)\colon z_{rc}\neq 1 \wedge
         \forall_{i\in R, j\in C, z_{ij}\neq 1 } ( x_{rc} \geq x_{ij})\}$ 
    
    \Else
    \State $M \gets \{(r,c)\colon \forall_{i\in R, j\in C} (x_{rc} \geq x_{ij})\}$ 
    \EndIf
    \State Randomly select $(r,c)$ from $M$ 
    \State $R\gets R \setminus \{r\}$ \Comment {Update the sets $R$ and $C$}
    \State $C\gets C \setminus \{c\}$
    \For {$i$ in $(1,n)$} 
        \State $x_{ri} \gets 0$ \Comment Clear $r$-th row
        \State $x_{ic} \gets 0$ \Comment Clear $c$-th column
	\EndFor 
    \State $x_{rc} \gets 1$  \Comment Assign 1 to the maximum element
	\EndFor
	\State return $X$
\EndProcedure
\end{algorithmic}
\end{algorithm}

\subsection{Migration}
\label{sub:migration}

The intended goal of the migration mechanism is to improve the algorithm exploration capabilities   by exchanging information between swarms. Actually, it is not a true migration, as particles do not move. Instead we modify stored $P^G[k]$ solutions (global best solution for a $k$-th swarm) replacing it by randomly picked solution from a swarm that performed better (see Algortithm~\ref{alg:migration}). 

The newly set $P^G[k]$ value influences the velocity vector for all particles in $k$-th swarm according to the formula (\ref{eq:velocity}). It may happen that the goal function value corresponding to the assigned solution $P^G[k])$ is worse than the previous one. It is accepted, as the migration is primarily designed  
to increase diversity within swarms.  

\begin{algorithm}
\caption{Migration procedure }
\label{alg:migration}
\begin{algorithmic}[1]
\Require $d$ - migration depth satisfying $d<m/2$, where $m$ is the number of swarms     
\Require $P^G$ - table of best solutions for $m$ swarms
\Require $X$ - set of all solutions

\Procedure{migration}{$d$,$P^G$,$X$}
    \State Sort swarms according to their $P^G_k$ values into \par a sequence $(s_1, s_2\dots,s_{m-2}, s_{m-1})$ 
    \For {$k$ in (1, $d$)}
    \State   Randomly choose a solution $x_{kj}$ \par \hskip 0.5cm belonging to the swarm $s_k$ 
	\State Assign $P^G[s_{m-k-1}] \gets x_{kj}$
	\State Update the best goal function value for \par \hskip 0.5cm the swarm $m-k-1$  
	\EndFor
\EndProcedure
\end{algorithmic}
\end{algorithm}

It should be mentioned that a naive approach consisting in copying best $P^G$ values between the swarms would be incorrect. (Consider replacing line 5 of Algorithm~\ref{alg:migration} with: $P^G[s_{m-k-1}] \gets P^G[s_{k}]$.) In such case during algorithm stagnation spanning over several iterations: in the first iteration the best value $P^G[1]$ would be cloned, in the second two copies would be created, in the third four and so on. Finally, after $k$ iterations $2^k$ swarms would follow the same direction. In the first group of experiments reported in Section~\ref{sec:experiments-results} we used up to 250 swarms. It means that after 8 iterations all swarms would be dominated by a single solution.     

\subsection{OpenCL algorithm implementation}

OpenCL \cite{khronos-openCL20} is a standard providing a common language, programming interfaces and hardware abstraction for heterogeneous platforms including GPU, multicore CPU, DSP and FPGA \cite{stone2010opencl}. It allows to accelerate computations by decomposing them into a set of parallel tasks (work items) operating on separate data. 

A program on OpenCL platform is decomposed into two parts: sequential executed by the CPU \emph{host} and parallel executed by multicore \emph{devices}. Functions executed on devices are called \emph{kernels}. They are written in a language being a variant of C with some restrictions related to keywords and datatypes. When first time loaded,  the kernels are automatically translated into the instruction set of the target device. The whole process takes about 500ms.

OpenCL supports 1D, 2D or 3D organization of data (arrays, matrices and volumes).
Each data element is identified by 1 to 3 indices, e.g. $d[i][j]$ for two-dimensional arrays. 
A \emph{work item} is a scheduled kernel instance, which obtains a combination of data indexes within the data range. 
To give an example, a 2D array of data of $n \times m$ size should be processed by $n\cdot m$ kernel instances, which are assigned with a pair of indexes $(i,j)$, $0 \le i<n$ and $0 \le j<m$. Those indexes are used to identify data items assigned to kernels.

Additionally, kernels can be organized into workgroups, e.g. corresponding to parts of a matrix, and synchronize their operations within a group using so called \emph{local barrier} mechanism.  However, workgroups suffer from several platform restrictions related to number of work items and amount of accessible memory.     

OpenCL uses three types of memory: global (that is exchanged between the host and the device), local for a work group and private for a work item. 

In our implementation we used \emph{aparapi}  platform \cite{aparapi} that allows to write OpenCL programs directly in Java language. The platform comprises two parts: an API and a runtime capable of converting Java bytecodes into OpenCL workloads. Hence, the host part of the program is executed on a Java virtual machine, and originally written in Java kernels are executed on an OpenCL enabled device.  

\begin{figure}
\centering
\includegraphics[width=0.45\columnwidth]{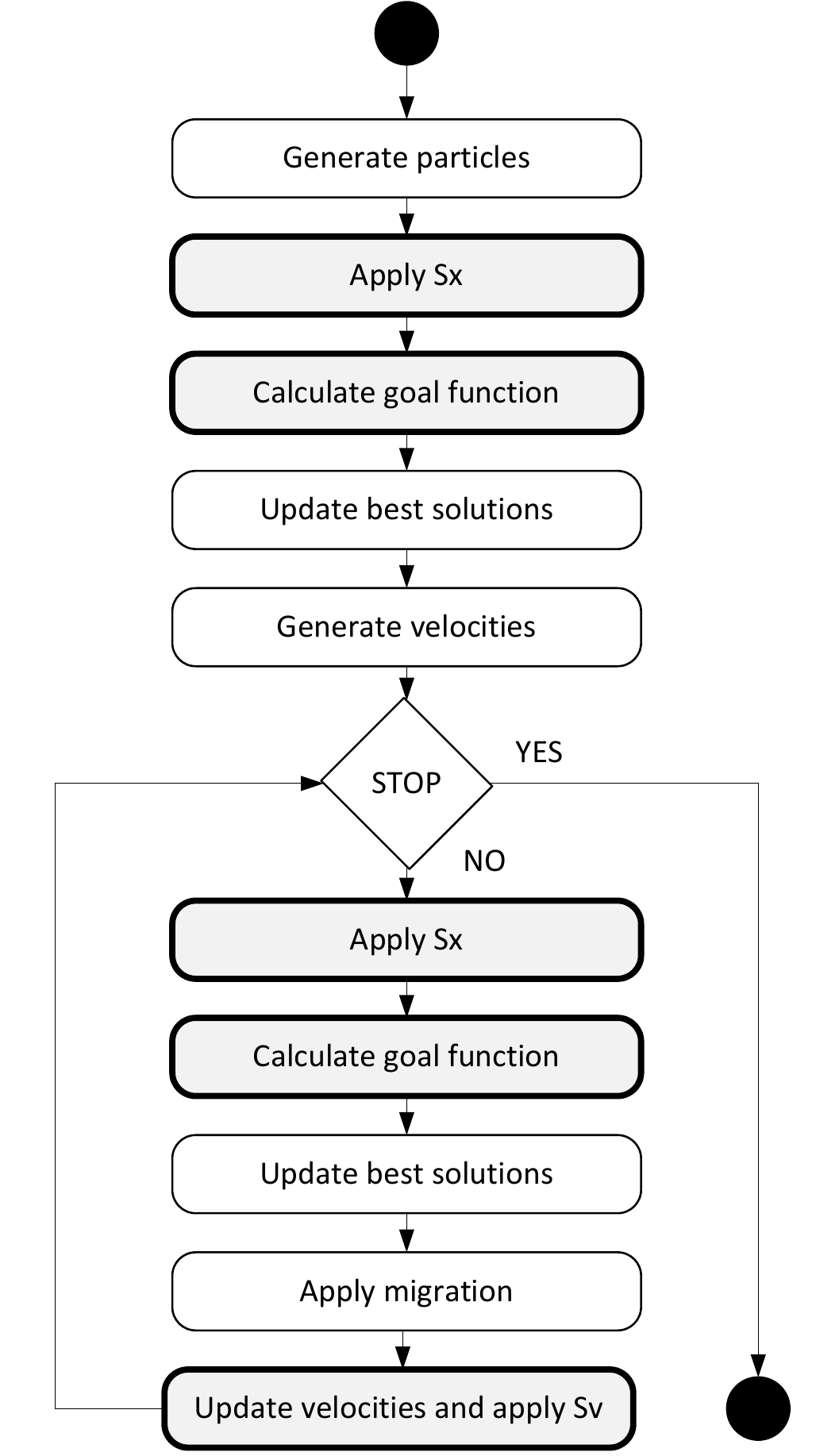}
\caption{Functional blocks of OpenCL based algorithm implementation. }
\label{fig:pso-qap-gpu-algorithm}
\end{figure}

The basic functional blocks of the algorithm are presented in Fig.~\ref{fig:pso-qap-gpu-algorithm}. Implemented kernels are marked with gray color. The code responsible for generation of random particles is executed by the host. We have also decided to leave the code for updating best solutions at the host side. Actually, it comprises a number of native {\tt System.arraycopy()} calls.  This  regards also the migration procedure, which sorts swarms indexes in a table according to their $P^G$ value and copies randomly picked entries.

\begin{figure}
\centering
\includegraphics[width=0.5\columnwidth]{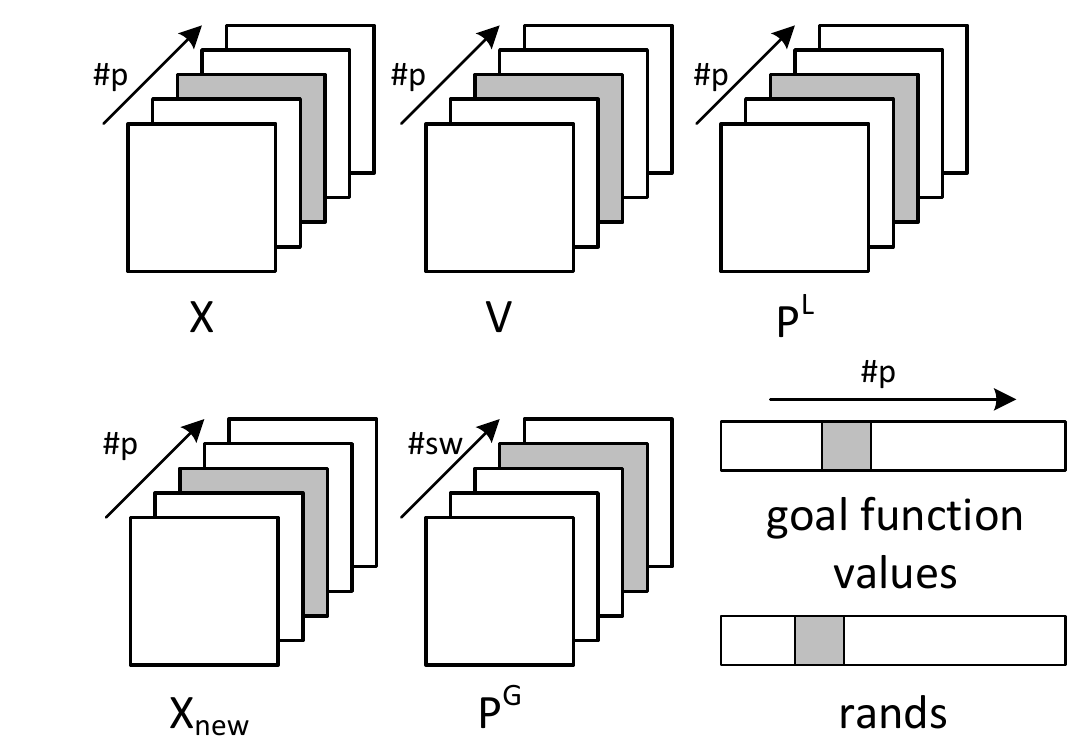}
\caption{Global variables used in the algorithm implementation}
\label{fig:pso-qap-global-variables}
\end{figure}

Particle data comprise  a number of matrices (see Fig.~\ref{fig:pso-qap-global-variables}): $X$ and $X_{new}$ -- solutions, $P^L$ -- local best particle solution and $V$ -- velocity. They are all stored in large continuous tables shared by all particles. Appropriate table part belonging to a particle  can be identified based on the \emph{particle id} transferred to a kernel. Moreover, while updating velocity, the particles reference a table  $P^G$ indexed by the \emph{swarm id}. 


An important decision related to OpenCL program design is related to data ranges selection. The memory layout in Fig.~\ref{fig:pso-qap-global-variables} suggests 3D range, whose dimensions are: row, column and particle number. This can be applied for relatively simple velocity or goal function calculation. However, the proposed algorithms for $S_x$, see Algorithm~\ref{alg:secondtarget}, are far too complicated to be implemented as a simple parallel work item. 
Finally we decided to use one dimension (particle id) for $S_x$ and goal function calculation, and two dimensions (particle id, swarm id) for velocity kernels.

%
%
%
%
%
%

It should be mentioned that requirements of parallel processing limits applicability of object oriented design at the host side. In particular we avoid creating particles or swarms with their own memory and then copying small chunks between the host and the device. Instead we rather use a flyweight design pattern \cite{gamma1994design}: if a  particle abstraction is needed, a single object can be configured to see parts of large global arrays $X$, $V$ as its own memory and perform required operations, e.g. initialization with random values.

\section{Experiments and results}
\label{sec:experiments-results}

In this section we report results of conducted experiments, which aimed at establishing the optimization performance of the implemented algorithm, as well as to collect data related to its statistical properties.

\subsection{Optimization results}

The algorithm was tested on several problem instances form the QAPLIB \cite{QAPLIB}, whose size ranged between 12 and 150. Their results are gathered in Table~\ref{tab:optimization-results} and Table~\ref{tab:brutal-force-results}. The selection of algorithm configuration parameters ($c_1$, $c_2$ and $c_3$ factors, as well as the kernels used) was based on previous results published in \cite{SzChKad2015}. In all cases the second target $S_x$ aggregation kernel was applied (see Algorithm~ \ref{alg:secondtarget}), which in previous experiments occurred the most successful.

During all tests reported in Table~\ref{tab:optimization-results}, apart the last, the total numbers of particles were large: 10000-12500. For the last case only 2500 particles were used due to 1GB memory limit of the GPU device (AMD Radeon HD 6750M card). In this case the consumed GPU memory ranged about 950~MB.

\begin{sidewaystable}
\caption{Results of tests for various instance from QAPLIB } 
\label{tab:optimization-results}
\vspace{12pt}
\centering
{ 

\begin{tabular}{|c|c|c|c|c|c|c|c|c||c|c|r|r|r|r|}
\hline
\hline
\rotatebox{-90}{No\hspace{6pt}}
&\rotatebox{-90}{Instance\hspace{6pt}}	
&\rotatebox{-90}{Size\hspace{6pt}}	
&\rotatebox{-90}{Number of swarms\hspace{6pt}}
&\rotatebox{-90}{Swarm size\hspace{6pt}}
&\rotatebox{-90}{Total particles\hspace{6pt}}
&\rotatebox{-90}{Inertia $c_1$\hspace{6pt}}
&\rotatebox{-90}{Self recognition $c_2$\hspace{6pt}}
&\rotatebox{-90}{Social factor $c_3$\hspace{6pt}}
&\rotatebox{-90}{Velocity kernel\hspace{6pt}}
&\rotatebox{-90}{Migration factor\hspace{6pt}}
&\rotatebox{-90}{Reached goal\hspace{6pt}}
&\rotatebox{-90}{Reference value\hspace{6pt}}
&\rotatebox{-90}{Gap\hspace{6pt}}
&\rotatebox{-90}{Iteration\hspace{6pt}}\\
\hline
\hline
1&chr12a	&12	&200	&50	&10000	&0.5	&0.5	&0.5	&Norm	&0	&9 552	&9 552	&0.00\%	&21\\
2&bur26a	&26	&250	&50	&12500	&0.8	&0.5	&0.5	&Raw	&33\%	&5 426 670	&5 426 670	&0.00\%	&156\\
3&bur26a	&26	&250	&50	&12500	&0.8	&0.5	&0.5	&Raw	&0	&5 429 693	&5 426 670	&0.06\%	&189\\
4&lipa50a	&50	&200	&50	&10000	&0.5	&0.5	&0.5	&Norm	&0	&62 794	&62 093	&1.13\%	&1640\\
5&tai60a	&60	&200	&50	&10000	&0.8	&0.5	&0.5	&Norm	&0	&7 539 614	&7 205 962	&4.63\%	&817\\
6&tai60a	&60	&300	&50	&15000	&0.8	&0.5	&0.5	&Raw	&0	&7 426 672	&7 205 962	&3.06\%	&917\\
7&tai60b	&60	&200	&50	&10000	&0.8	&0.3	&0.3	&Norm	&0	&620 557 952	&608 215 054	&2.03\%	&909\\
8&tai60b	&60	&200	&50	&10000	&0.5	&0.5	&0.5	&Norm	&0	&617 825 984	&608 215 054	&1.58\%	&1982\\
9&tai60b	&60	&200	&50	&10000	&0.8	&0.3	&0.3	&Norm	&33\%	&612 078 720	&608 215 054	&0.64\%	&2220\\
10&tai60b	&60	&200	&50	&10000	&0.8	&0.3	&0.3	&Norm	&33\%	&614 088 768	&608 215 054	&0.97\%	&1619\\
11&tai64c	&64	&200	&50	&10000	&0.8	&0.5	&0.5	&Norm	&0	&1 856 396	&1 855 928	&0.03\%	&228\\
12 &esc64a	&64	&200	&50	&10000	&0.8	&0.5	&0.5	&Raw	&0	&116	&116	&0.00\%	&71\\

13&tai80a	&80	&200	&50	&10000	&0.8	&0.5	&0.5	&Raw	&0	&14 038 392	&13 499 184	&3.99\%	&1718\\
14&tai80b	&80	&200	&50	&10000	&0.8	&0.5	&0.5	&Raw	&0	&835 426 944	&818 415 043	&2.08\%	&1509\\
15&sko100a	&100	&200	&50	&10000	&0.8	&0.5	&0.5	&Raw	&0	&154 874	&152 002	&1.89\%	&1877\\
16&tai100b	&100	&200	&50	&10000	&0.8	&0.5	&0.5	&Raw	&0	&1 196 819 712	&1 185 996 137	&0.91\%	&1980\\
17 &esc128	&128	&100	&50	&5000	&0.8	&0.5	&0.5	&Raw	&0	&64	&64	&0.00\%	&1875\\
18&tai150b	&150	&50	&50	&2500	&0.8	&0.5	&0.5	&Raw	&0	&530 816 224	&498 896 643	&6.40\%	&1894\\

\hline

\end{tabular}
}
\end{sidewaystable}

The results show that algorithm is capable of finding solutions with goal function values are close to reference numbers listed in QAPLIB. The gap is between 0\% and 6.4\% for the biggest case \emph{tai150b}. We have repeated tests for \emph{tai60b} problem to compare the implemented multi-swarm algorithm with the previous single-swarm version published in \cite{SzChKad2015}. Gap values for the best results obtained with the single swarm algorithm  were around  7\%-8\%. For the multi-swarm implementation discussed here the gaps were between 0.64\% and 2.03\%.

{

The goal of the second group of experiments was to test the algorithm configured to employ large numbers of particles (50000-10000) for well known \emph{esc32*} problem instances from the QAPLIB. Altough they were considered hard, all of them have been were recently solved optimally with exact algorithms \cite{nyberg2012new,Fischetti2012}. 

The results are summarized in Table~\ref{tab:brutal-force-results}.  We used the following parameters: $c_1=0.8$, $c_2=0.5$ $c_3=0.5$, velocity kernel: normalized, $S_x$ kernel: second target. 
During nearly all experiments optimal values of goal functions were reached in one algorithm run. Only the problem \emph{esc32a} occurred difficult,  therefore for this case the number of particles, as well as the upper iteration limits were increased to reach the optimal solution. What was somehow surprising, in all cases  solutions differing from those listed in QAPLIB were obtained. Unfortunately, our algorithm was not prepared to collect sets of optimal solutions, so we are not able to provide detailed results on their numbers.
 
It can be seen that optimal solutions for problem instances \emph{esc32c--h} were found in relatively small numbers of iterations. In particular, for \emph{esc32e} and \emph{es32g}, which are characterized by small values of goal functions, optimal solutions were found during the initialization or in the first iteration.

\begin{table}
\caption{Results of tests for \emph{esc32*} instances from QAPLIB (problem size $n=32$).  Reached optimal values are marked with asterisks. } 
\label{tab:brutal-force-results}
\centering
\vspace{12pt}
{
\begin{tabular}{|c|c|c|c|c|c|r|r|r|}
\hline
Instance&Swarms&Particles&Total part.&Goal&Iter&Time/iter [ms]\\
\hline
esc32a&50&1000&100000&138&412&3590.08\\
esc32a&10&5000&100000&134&909&3636.76\\
esc32a&50&2000&100000&$130^*$&2407&3653.88\\
esc32b&50&1000&50000&$168^*$&684&3637.84\\
esc32c&50&1000&50000&$642^*$&22&3695.19\\
esc32d&50&1000&50000&$400^*$&75&3675.32\\
esc32e&50&1000&50000&$2^*$&0&3670.38\\
esc32g&50&1000&50000&$6^*$&1&3625.17\\
esc32h&50&1000&50000&$438^*$&77&3625.17\\

\hline

\end{tabular}
}
\end{table}

}

The disadvantage of the presented algorithm is that it uses internally matrix representation for solutions and velocities. In consequence the memory consumption is proportional to $n^2$, where $n$ is the problem size. The same regards the time complexity, which for goal function and $S_x$ procedures can be estimated as $o(n^3)$. This makes optimization of large problems time consuming (e.g. even 400 sec for one iteration for \emph{tai150b}). However, for for medium size problem instances, the iteration times  are much smaller, in spite of large populations used. For two runs of the algorithm \emph{bur26a} reported in Table~\ref{tab:optimization-results}, where during each iteration  12500 particles were processed, the average iteration time was equal 1.73 sec. {
For 50000-10000 particles  and problems of size $n=32$ the average iteration time reported in Table~\ref{tab:brutal-force-results} was less than 3.7 seconds.}

\subsection{Statistical results}

{
An obvious benefit of massive parallel computations  is the capability of processing large populations (see Table~\ref{tab:brutal-force-results}). Such approach to optimization may resemble a little bit a \emph{brutal force} attack: the solution space is randomly sampled millions of times to hit the best solution. 
No doubt that such approach can be more successful if combined with a correctly designed exploration mechanism that directs the random search process towards solutions providing  good or near-optimal solutions.
In this section we analyze collected statistical data related to the algorithm execution to show that the optimization performance of the algorithm can be attributed not only to large sizes of processed population, but also to the implemented exploration mechanism.     
}

PSO algorithm can be considered a stochastic process controlled by random variables $r_2(t)$ and $r_3(t)$ appearing in its state equation (\ref{eq:pso-state-eq}). Such analysis for continuous problems were conducted in \cite{FernandezGarcia09}.
On the other hand, the observable algorithm outcomes, i.e. the values of goal functions $f(x_i(t))$ for solutions $x_i$, $i=1,\dots,n$ reached in consecutive time moments $t\in\{1,2,3,\dots\}$ can be also treated as random variables, whose distributions change over time $t$. Our intuition is that a correctly designed algorithm should result in a nonstationary   
stochastic process $\{f(x_i(t))\colon t\in T\}$, characterized by growing probability that next values of goal functions in the analyzed population are closer to the optimal solution.

To demonstrate such behavior of the implemented algorithm we have collected detailed information on goal function values during two optimization task for the problem instance \emph{bur26a} reported in Table~\ref{tab:optimization-results} (cases 2 and 3). For both of them the algorithm was configured to use 250 swarms comprising 50 particles. In the case 2 the migration mechanism was applied and the optimal solution was found in the iteration 156, in the case 3 (without migration) a solution very close to optimal (gap 0.06\%) was reached in the iteration 189.
       
\begin{figure}[!ht]
 \centering
  \includegraphics[width=0.7\columnwidth]{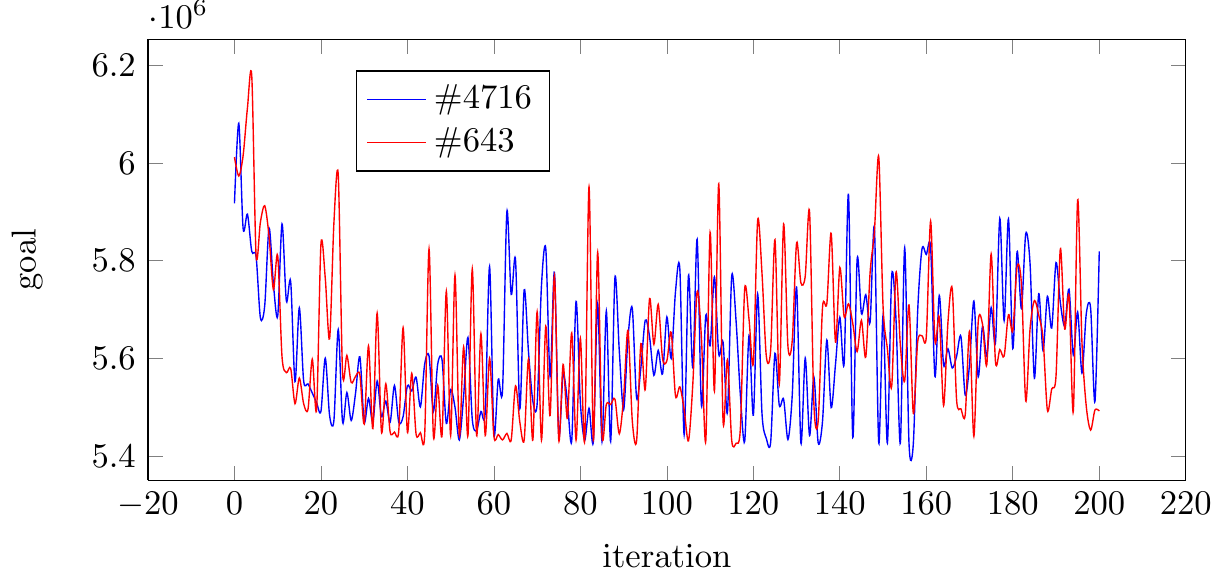}
  \caption{Variations of goal function values for two particles exploring the solutions space during the optimization process (bur26a problem instance) }
  \label{fig:particle-goal}
\end{figure}

Fig.~\ref{fig:particle-goal} shows values of goal function for two selected particles during run 3. The plots  show  typical QAP specificity. PSO and many other algorithms perform a local neighborhood search. 
For the QAP the neighborhood is characterized by great variations of goal function values. Although mean values of goal function decrease in first twenty or thirty iterations, the particles behave randomly and nothing indicates that during subsequent iterations smaller values of goal functions would be reached more often.

\begin{figure}[!ht]
\centering
  \includegraphics[width=0.7\columnwidth]{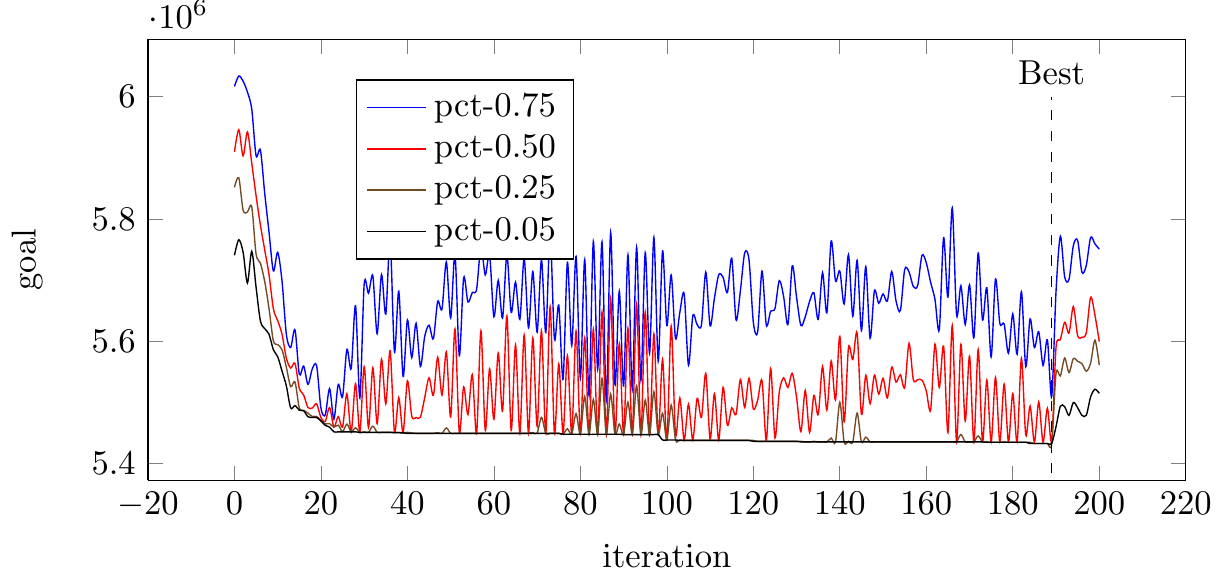}
  \includegraphics[width=0.7\columnwidth]{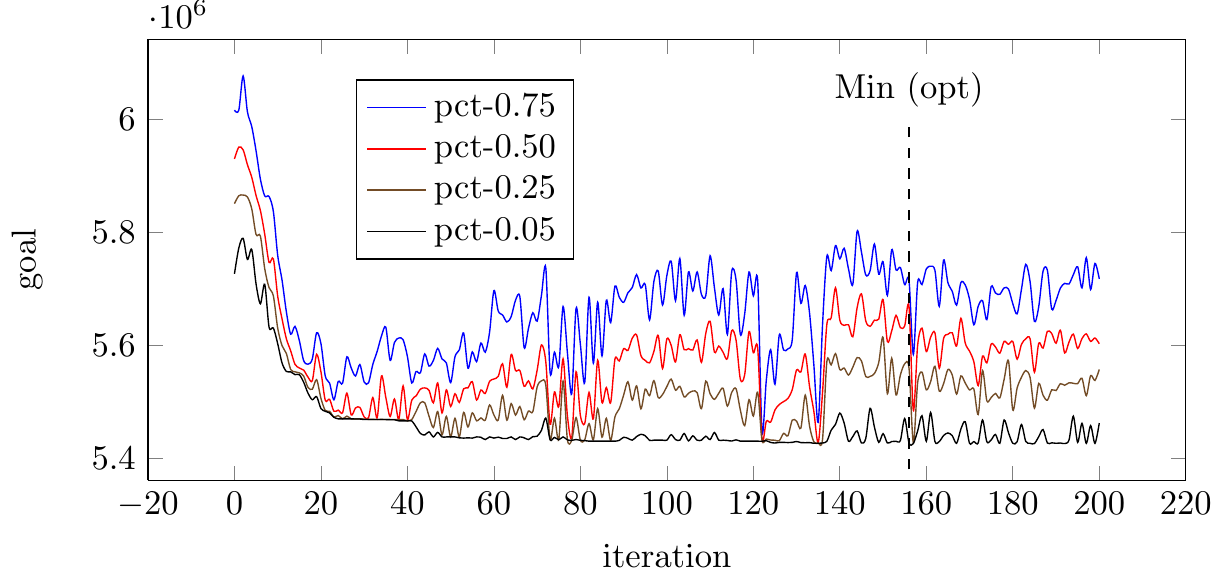}
\caption{Two runs of bur26a optimization. Percentile ranks for 50 particles belonging to the most successful swarms: without migration (above) and with migration (below).}    
\label{fig:swarm-percentile}
\end{figure}

In Fig.~\ref{fig:swarm-percentile} percentile ranks (75\%, 50\% 25\% and 5\%) for two swarms, which reached best values in cases 2 and 3 are presented. Although the case 3 is characterized by less frequent changes of scores, than the case 2, probably this effect can not be attributed to the migration applied. It should be mentioned that for a swarm comprising 50 particles, the 0.05 percentile corresponds to just two of them.    

\begin{figure}[!ht]
\centering
  \includegraphics[width=0.7\columnwidth]{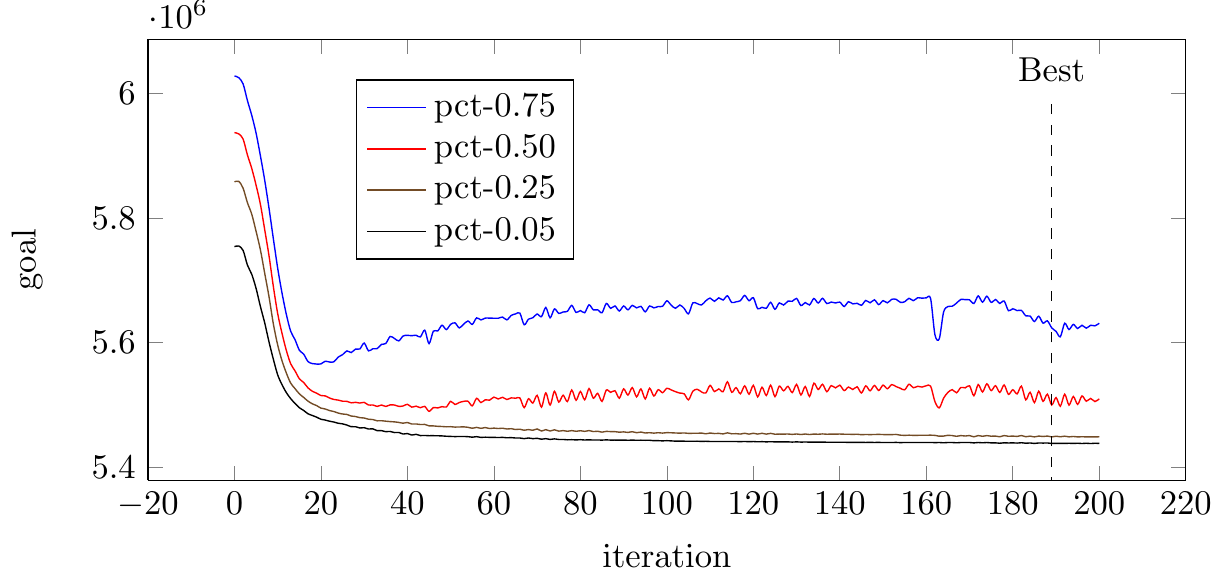}
  \includegraphics[width=0.7\columnwidth]{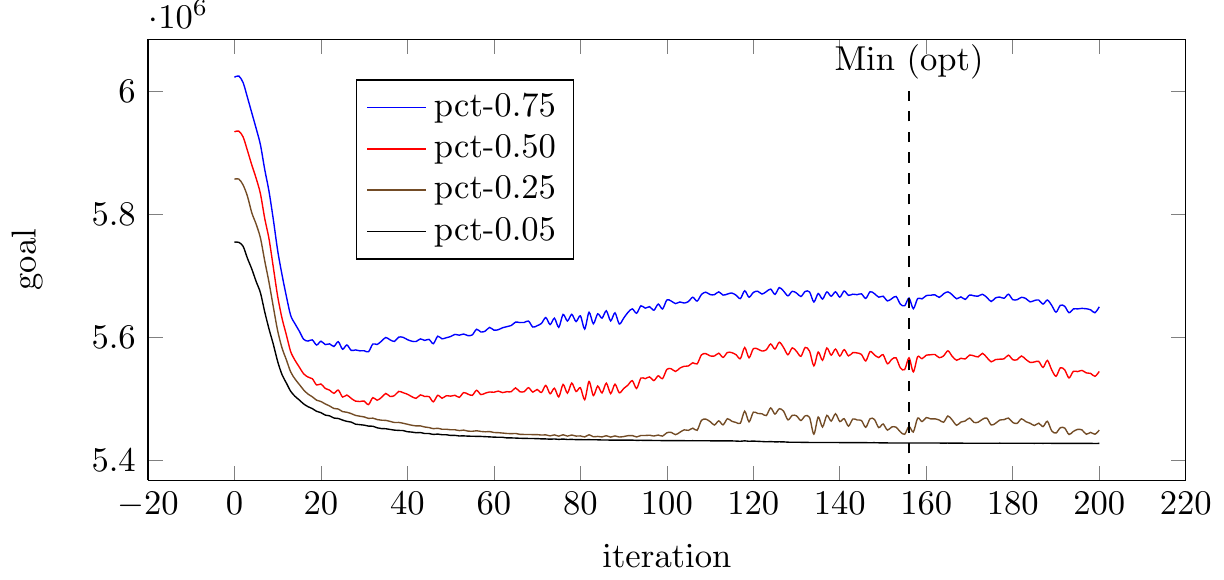}
\caption{Two runs of bur26a optimization. Percentile ranks for all 12500 particles: without migration (above) and with migration (below). }
\label{fig:global-percentile}
\end{figure}

Collected percentile rank values for the whole population comprising 12500 particles are presented in  Fig.~\ref{fig:global-percentile}. For both cases the plots are clearly separated. It can be also observed that solutions very close to optimal are practically reached between the iterations 20 (37.3 sec) and 40 (72.4 sec).  For the whole population the 0.05 percentile represents 625 particles. Starting with the iteration 42 their score varies between $5.449048\cdot 10^6$ and $5.432361\cdot 10^6$, i.e. by about $0.3\%$.

\begin{figure}[!ht]
\centering
  \includegraphics[width=0.7\columnwidth]{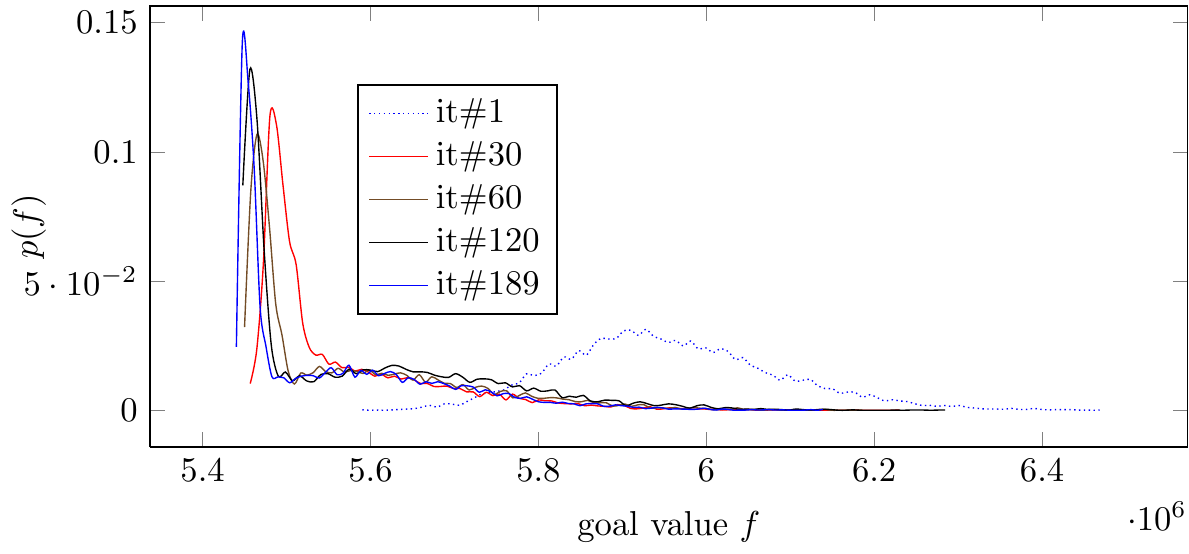}
  \includegraphics[width=0.7\columnwidth]{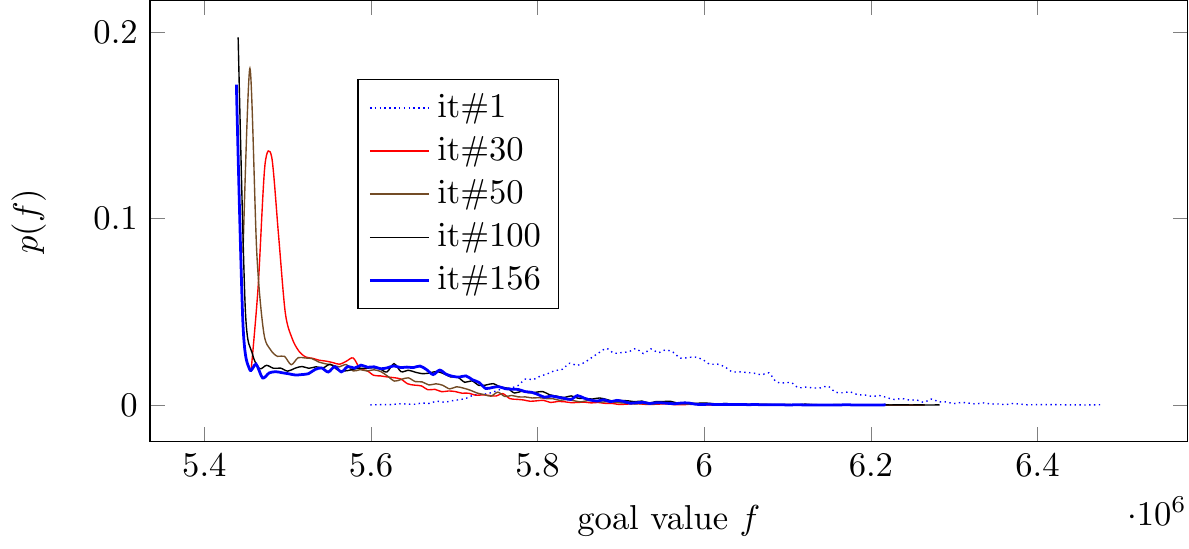}
\caption{Probability mass functions  for 12050 particles organized into 250 x 50  swarms during two runs: without migration (above) and with migration (below).  }
\label{fig:global-prob-distribution}
\end{figure}

Fig.~\ref{fig:global-prob-distribution} shows, how the probability distribution (probability mass function PMF) changed during the optimization process. In both cases the the optimization process starts with a normal distribution with the mean value about 594500. In the subsequent iterations the maximum of PMF grows and moves towards smaller values of the goal function. There is no fundamental difference between the two cases, however for the case 3 (with migration) maximal values of PMF are higher. It can be also observed that in the iteration 30 (completed in 56 seconds) the probability of hitting a good solution is quite high, more then 10\%.          

\begin{figure}[!ht]
\centering
  \includegraphics[width=0.7\columnwidth]{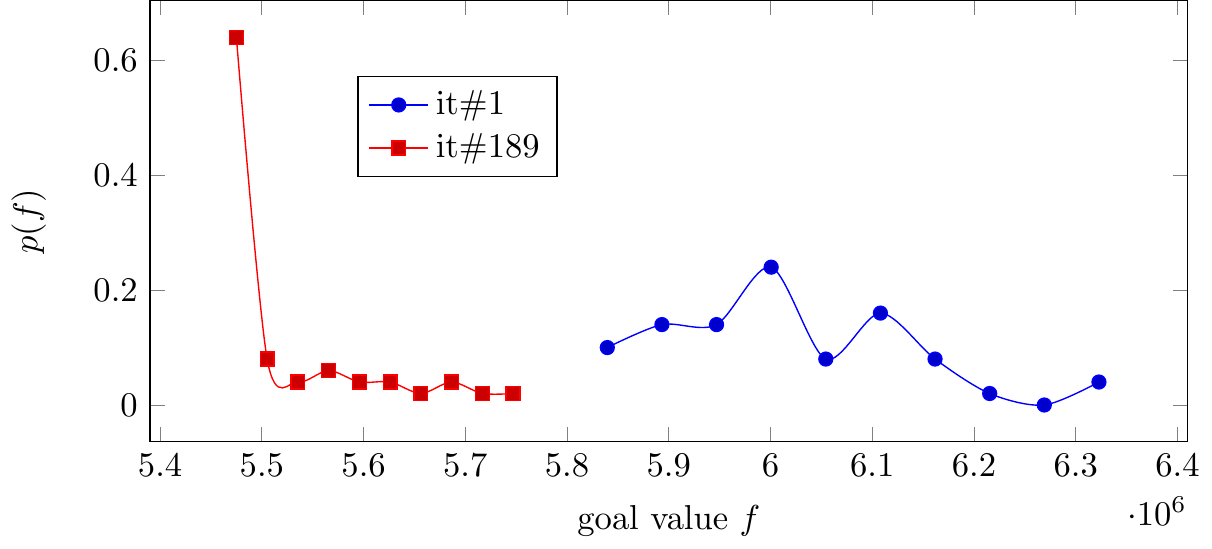}
  \includegraphics[width=0.7\columnwidth]{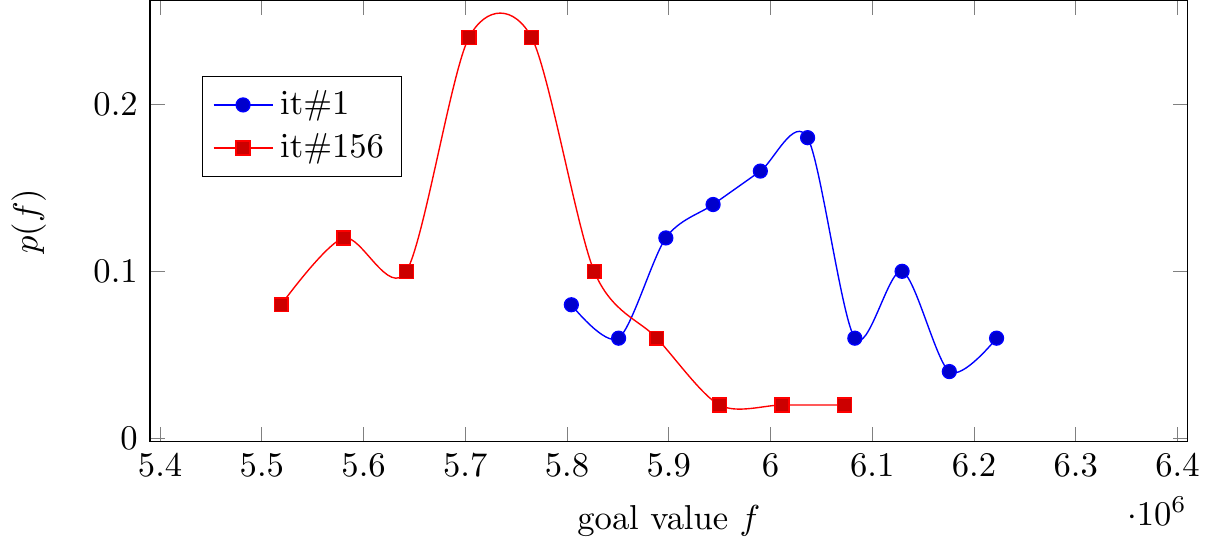}
\caption{Probability mass functions for 50 particles belonging to the most successful swarms during two runs: without migration (above) and with migration (below). One point represents an upper bound for 5 particles.}
\label{fig:swarm-prob-distibution}
\end{figure}

Interpretation of PMF for the two most successful swarms that reached best values in the discussed cases is not that obvious. For the case without migration (Fig.~\ref{fig:swarm-prob-distibution} above) there is a clear separation between the initial distribution and the distribution reached in the iteration, which yielded the best result. In the second case (with migration) a number of particles were concentrated around local minima.

The presented data shows advantages of optimization performed on massive parallel processing platforms. Due to high number of solutions analyzed simultaneously, the algorithm that does not exploit the problem structure can yield acceptable results in relatively small number of iterations (and time). For a low-end GPU devices, which was used during the test, \emph{good enough} results were obtained after 56 seconds.   
It should be mentioned that for both presented cases the maximum number of iterations was set to 200. With 12500 particles, the ratio of potentially explored solutions to the whole solution space was equal $200\cdot 12500/26!=6.2\cdot 10^{-21}$.

\section{Conclusions}
\label{sec:conclusions}

In this paper we describe a multi-swarm PSO algorithm for solving the QAP problem designed for 
the OpenCL platform. The algorithm is capable of processing in parallel large number of particles organized into several swarms that either run independently or communicate with use of the migration mechanism. 
Several solutions related to particle state representation and particle movement  were inspired by the work of Liu at al.  \cite{Liu2007}, however, they were refined here to provide better performance.  

We tested the algorithm on several problem instances from the QAPLIB library obtaining good results (small gaps between reached solutions and reference values). However, it seems that for problem instances of large sizes the selected representation of solutions in form of permutation matrices hinders potential benefits of parallel processing.  

During the experiments the algorithm was configured to process large populations. This allowed us to collect statistical data related to goal function values reached by individual particles. We used them to demonstrate on two cases that although single particles seem to behave chaotically during the optimization process, when the whole population is analyzed, the probability that a particle will select a near-optimal solution grows. This growth is significant for a number of initial iterations, then its speed diminishes and finally reaches zero. 

Statistical analysis of experimental data collected during optimization process may help to tune the algorithm parameters, as well as to establish realistic limits related to expected improvement of goal functions.       
This in particular regards practical applications of optimization techniques, in which recurring optimization problems appear, i.e. the problems with similar size, complexity and structure. 
Such problems can be near-optimally solved in bounded time on massive parallel computation platforms even, if low-end devices are used.


%
%
\bibliographystyle{splncs03}

\bibliography{wch,pszwed,qap,pso,opencl,cuda}

\begin{thebibliography}{10}
\providecommand{\url}[1]{\texttt{#1}}
\providecommand{\urlprefix}{URL }

\bibitem{ahuja2000greedy}
Ahuja, R.K., Orlin, J.B., Tiwari, A.: A greedy genetic algorithm for the
  quadratic assignment problem. Computers \& Operations Research  27(10),
  917--934 (2000)

\bibitem{anstreicher2002solving}
Anstreicher, K., Brixius, N., Goux, J.P., Linderoth, J.: Solving large
  quadratic assignment problems on computational grids. Mathematical
  Programming  91(3),  563--588 (2002)

\bibitem{stickel:Bermudez01}
Bermudez, R., Cole, M.H.: A genetic algorithm approach to door assignments in
  breakbulk terminals. Tech. Rep. MBTC-1102, Mack-Blackwell Transportation
  Center, University of Arkansas, Fayetteville, Arkansas (2001)

\bibitem{BKR91}
Burkard, R.E., Karisch, S.E., Rendl, F.: {QAPLIB} - a {Q}uadratic {A}ssignment
  {P}roblem library. Journal of Global Optimization  10(4),  391--403 (1997)

\bibitem{Cela98}
\c{C}ela, E.: {The quadratic assignment problem: theory and algorithms}.
  Combinatorial Optimization, Springer, Boston (1998)

\bibitem{ChK11}
Chmiel, W., Kad{\l}uczka, P., Packanik, G.: Performance of swarm algorithms for
  permutation problems. Automatyka  15(2),  117--126 (2009)

\bibitem{clerc2004discrete}
Clerc, M.: Discrete particle swarm optimization, illustrated by the traveling
  salesman problem. In: New optimization techniques in engineering, pp.
  219--239. Springer (2004)

\bibitem{EberhartKennedy95}
Eberhart, R., Kennedy, J.: A new optimizer using particle swarm theory. In:
  Micro Machine and Human Science, 1995. MHS '95., Proceedings of the Sixth
  International Symposium on. pp. 39--43 (Oct 1995)

\bibitem{FernandezGarcia09}
Fernández~Martínez, J., García~Gonzalo, E.: The {PSO} family: deduction,
  stochastic analysis and comparison. Swarm Intelligence  3(4),  245--273
  (2009)

\bibitem{Fischetti2012}
Fischetti, M., Monaci, M., Salvagnin, D.: Three ideas for the quadratic
  assignment problem. Operations Research  60(4),  954--964 (2012)

\bibitem{fon2010investigating}
Fon, C.W., Wong, K.Y.: Investigating the performance of bees algorithm in
  solving quadratic assignment problems. International Journal of Operational
  Research  9(3),  241--257 (2010)

\bibitem{gambardella1999ant}
Gambardella, L.M., Taillard, E., Dorigo, M.: Ant colonies for the quadratic
  assignment problem. Journal of the operational research society pp. 167--176
  (1999)

\bibitem{gamma1994design}
Gamma, E., Helm, R., Johnson, R., Vlissides, J.: Design Patterns: Elements of
  Reusable Object-Oriented Software. Pearson Education (1994)

\bibitem{Groetschel1991}
Gr{\"o}tschel, M.: Discrete mathematics in manufacturing. In: Malley, R.E.O.
  (ed.) ICIAM 1991: Proceedings of the Second International Conference on
  Industrial and Applied Mathematics. pp. 119--145. SIAM (1991)

\bibitem{hahn2013memory}
Hahn, P., Roth, A., Saltzman, M., Guignard, M.: Memory-aware parallelized rlt3
  for solving quadratic assignment problems. Optimization online  (2013),
  \url{http://www.optimization-online.org/DB_HTML/2013/12/4144.html}

\bibitem{hahn2010exact}
Hahn, P.M., Zhu, Y.R., Guignard, M., Smith, J.M.: Exact solution of emerging
  quadratic assignment problems. International Transactions in Operational
  Research  17(5),  525--552 (2010)

\bibitem{aparapi}
Howes, L., Munshi, A.: Aparapi - {AMD}.
  \url{http://developer.amd.com/tools-and-sdks/opencl-zone/aparapi/}, online:
  last accessed: Jan 2015

\bibitem{khronos-openCL20}
Howes, L., Munshi, A.: The {OpenCL} specification.
  \url{https://www.khronos.org/registry/cl/specs/opencl-2.0.pdf}, online: last
  accessed: Jan 2015

\bibitem{KBe57}
Koopmans, T.C., Beckmann, M.J.: Assignment problems and the location of
  economic activities. Econometrica  25,  53--76 (1957)

\bibitem{memeticGPU2013}
Krüger, F., Maitre, O., Jiménez, S., Baumes, L., Collet, P.: Generic local
  search (memetic) algorithm on a single {GPGPU} chip. In: Tsutsui, S., Collet,
  P. (eds.) Massively Parallel Evolutionary Computation on {GPGPUs}, pp.
  63--81. Natural Computing Series, Springer Berlin Heidelberg (2013)

\bibitem{Liu2007}
Liu, H., Abraham, A., Zhang, J.: A particle swarm approach to quadratic
  assignment problems. In: Saad, A., Dahal, K., Sarfraz, M., Roy, R. (eds.)
  Soft Computing in Industrial Applications, Advances in Soft Computing,
  vol.~39, pp. 213--222. Springer Berlin Heidelberg (2007)

\bibitem{geneticGPU2013}
Maitre, O.: Genetic programming on {GPGPU} cards using {EASEA}. In: Tsutsui,
  S., Collet, P. (eds.) Massively Parallel Evolutionary Computation on
  {GPGPUs}, pp. 227--248. Natural Computing Series, Springer Berlin Heidelberg
  (2013)

\bibitem{MasonR97}
Mason, A., R{\"o}nnqvist, M.: Solution methods for the balancing of jet
  turbines. Computers \& OR  24(2),  153--167 (1997)

\bibitem{Misevicius12}
Misevicius, A.: An implementation of the iterated tabu search algorithm for the
  quadratic assignment problem. OR Spectrum  34(3),  665--690 (2012)

\bibitem{nyberg2012new}
Nyberg, A., Westerlund, T.: A new exact discrete linear reformulation of the
  quadratic assignment problem. European Journal of Operational Research
  220(2),  314--319 (2012)

\bibitem{onwubolu2004particle}
Onwubolu, G.C., Sharma, A.: Particle swarm optimization for the assignment of
  facilities to locations. In: New Optimization Techniques in Engineering, pp.
  567--584. Springer (2004)

\bibitem{owens2007survey}
Owens, J.D., Luebke, D., Govindaraju, N., Harris, M., Kr{\"u}ger, J., Lefohn,
  A.E., Purcell, T.J.: A survey of general-purpose computation on graphics
  hardware. In: Computer graphics forum. vol.~26, pp. 80--113. Wiley Online
  Library (2007)

\bibitem{QAPLIB}
{Peter Hahn and Miguel Anjos}: {QAPLIB} home page.
  \url{http://anjos.mgi.polymtl.ca/qaplib/}, online: last accessed: Jan 2015

\bibitem{Phillips94aquadratic}
Phillips, A.T., Rosen, J.B.: A quadratic assignment formulation of the
  molecular conformation problem. JOURNAL OF GLOBAL OPTIMIZATION  4,  229--241
  (1994)

\bibitem{Sahni76}
Sahni, S., Gonzalez, T.: P-complete approximation problems. J. ACM  23(3),
  555--565 (1976)

\bibitem{stone2010opencl}
Stone, J.E., Gohara, D., Shi, G.: Opencl: A parallel programming standard for
  heterogeneous computing systems. Computing in science \& engineering  12(3),
  ~66 (2010)

\bibitem{stutzle1999aco}
St{\"u}tzle, T., Dorigo, M.: Aco algorithms for the quadratic assignment
  problem. New ideas in optimization pp. 33--50 (1999)

\bibitem{SzChKad2015}
Szwed, P., Chmiel, W., Kad{\l}uczka, P.: {OpenCL} implementation of {PSO}
  algorithm for the {Quadratic Assignment Problem}. In: Rutkowski, L.,
  Korytkowski, M., Scherer, R., Tadeusiewicz, R., Zadeh, L.A., Zurada, J.M.
  (eds.) Artificial Intelligence and Soft Computing, Lecture Notes in Computer
  Science, vol. Accepted for ICAISC'2015 Conference. Springer International
  Publishing (2015),
  \url{http://home.agh.edu.pl/~pszwed/en/lib/exe/fetch.php?media=papers:draft-icaics-2015-pso-qap-opencl.pdf}

\bibitem{Taillard199587}
Taillard, E.D.: Comparison of iterative searches for the quadratic assignment
  problem. Location Science  3(2),  87 -- 105 (1995)

\bibitem{acoTabuGPU2013}
Tsutsui, S., Fujimoto, N.: {ACO} with tabu search on {GPUs} for fast solution
  of the {QAP}. In: Tsutsui, S., Collet, P. (eds.) Massively Parallel
  Evolutionary Computation on {GPGPUs}, pp. 179--202. Natural Computing Series,
  Springer Berlin Heidelberg (2013)

\bibitem{zhou2009gpu}
Zhou, Y., Tan, Y.: {GPU}-based parallel particle swarm optimization. In:
  Evolutionary Computation, 2009. CEC'09. IEEE Congress on. pp. 1493--1500.
  IEEE (2009)

\end{thebibliography}

\label{lastpage}
\end{document}